%% file: iccv_main.tex
\documentclass[10pt,twocolumn,letterpaper]{article}

\usepackage{iccv}
\usepackage{times}
\usepackage{epsfig}
\usepackage{graphicx}
\usepackage{amsmath}
\usepackage{amssymb}
\usepackage{bbm}
\usepackage{multirow}

\usepackage{bbding}
\usepackage{graphicx}
\usepackage{graphicx}
\usepackage{amsmath}
\usepackage{amssymb}
\usepackage{booktabs}
\usepackage{bbm}
\usepackage{colortbl}
\usepackage{flushend}
\usepackage[dvipsnames]{xcolor}
\usepackage{bm}
\usepackage{multirow}
\usepackage{makecell}
\usepackage{tikz}
\usepackage{comment}
\usepackage{amsmath,amssymb} 
\usepackage{color}

\usepackage[pagebackref=true,breaklinks=true,letterpaper=true,colorlinks,bookmarks=false]{hyperref}

\iccvfinalcopy 

\def\iccvPaperID{5804} 

\ificcvfinal\fi

\usepackage[capitalize]{cleveref}
\usepackage{expl3}
\ExplSyntaxOn
\newcommand\latinabbrev[1]{
	\peek_meaning:NTF . {
		#1\@}%
	{ \peek_catcode:NTF a {
			#1.\@ }%
		{#1.\@}}} 
\ExplSyntaxOff

\def\eg{\latinabbrev{e.g}}

\def\ie{\latinabbrev{i.e}}

\DeclareMathOperator*{\argmax}{arg\,max}

\newcommand{\rrvos}{R\textsuperscript{2}-VOS}
\newcommand{\rrytbvos}{R\textsuperscript{2}-Youtube-VOS}

\begin{document}

\title{Robust  Referring Video Object Segmentation with Cyclic Structural Consensus}

\author{Xiang Li$^{1}$\thanks{This work was done when Xiang Li and Xiaohao Xu were interns at MSRA.}, Jinglu Wang$^2$, Xiaohao Xu$^{3}$, Xiao Li$^2$, Bhiksha Raj$^{1,4}$, Yan Lu$^2$ \\
$^1$Carnegie Mellon University, $^2$Microsoft Research Asia, $^3$University of Michigan,\\ $^4$Mohamed bin Zayed University of Artificial Intelligence\\
}

\maketitle
\ificcvfinal\thispagestyle{empty}\fi

\input{src/0-abstract}
\input{src/1-introduction}

\input{src/2-related_works}
\input{src/3-problem}

\input{src/4-method}
\input{src/5-experiments}

\input{src/6-conclusion}
\clearpage
{\small
\bibliographystyle{ieee_fullname}
\bibliography{src/reference}
}


\end{document}


\title{Robust  Referring Video Object Segmentation with Cyclic Structural Consensus}

\author{First Author\\
Institution1\\
Institution1 address\\
{\tt\small firstauthor@i1.org}
\and
Second Author\\
Institution2\\
First line of institution2 address\\
{\tt\small secondauthor@i2.org}
}

\maketitle
\ificcvfinal\thispagestyle{empty}\fi

\section{Dataset Generation}
In this section, we elaborate on the details of $\mathrm{R}^2$-VOS dataset generation. To comprehensively evaluate the robustness quality, we generate the negative expressions with different methods as follows.  
\begin{itemize}
    \item \textbf{Replacing properties.} We replace one or more of the \{category, action, appearance, position\} properties in the original expression to construct the negative expressions.
    \item \textbf{Replacing entire expression with other expression in the dataset.} We replace the entire original expression with other expressions in Ref-Youtube-VOS dataset to construct negative expressions. 
    \item \textbf{Replacing entire expression with other expression out of the dataset.} We replace the entire original expression with expressions in other dataset (Ref-COCO) to construct negative expressions. 
\end{itemize}
In all, we generate 4982 negative samples (about 10 negative expressions per video). For the negative expressions generated by replacing the original expression, we first leverage object category as a criterion to check whether a new object can be referred by the negative expression. For the videos containing the object of the category described by the expression, we manually check those cases to ensure the negative expression cannot correspond to any object in the video. 

\section{More Quantitative Result Analysis}
 Under the same ResNet-50 backbone, our method achieves 57.3 $\mathcal{J}\&\mathcal{F}$, 94.1 $\mathcal{R}$ and 30 FPS compared to the 55.6 $\mathcal{J}\&\mathcal{F}$, 30.6 $\mathcal{R}$ and 22 FPS of ReferFormer. We will then point-to-point analyze reasons of improvements on $\mathcal{J}\&\mathcal{F}$ (for positive video), $\mathcal{R}$ (for negative videos) and FPS (for inference speed). 
\begin{itemize}
    \item $\mathcal{J}\&\mathcal{F}$: (1) We introduce the early-grounding module which employs both pixel-wise and channel-wise attention to enable multimodal interaction. Different from the CM-FPN used in ReferFormer that solely fuses features from text to video in pixel-level, our early-grounding module first enables pixel-level bi-directional fusion and then generates dynamic kernels using the fused text feature $\mathrm{\mathbf{g^\prime}}$ to modulate the video feature $\mathrm{\mathbf{f^\prime}}$. The dynamic convolution (channel-wise attention) is commonly used to decode dense masks from visual features and is suitable to suppress irrelevant features. By equipping text-guided dynamic convolution in early-stage, the pixel decoder can be more focused on the target object (as shown in Figure~4). (2) Our method leverages relational cycle consistency to constraint the intermediate feature $f_{early}$ to contain correct object-level information to recover some properties of original text embedding. By applying this constraint, our method can better avoid interference and easier locate the correct object. (3) Our instance query is composed of both original sentence embedding and the reconstructed one. Different from ReferFormer that only utilizes original sentence embedding as queries, the reconstructed embedding can encode visual information to facilitate the instance query decode the objects from visual features.
    \item $\mathcal{R}$: The newly introduced metric $\mathcal{R}$ aims to measure the robustness of the model against unpaired inputs. Text-video pairs with (object-level) semantic consensus can be assumed as in-distribution for RVOS problem where semantic consensus can be kind of easily modeled. In contrast, unpaired text-video is much more difficult to tackle because there can be unlimited out-of-distribution (OOD) scenarios for the text-video pairs. In our method, instead of directly detect the OOD of input pairs, we convert the problem to find semantic alignment between the input text embedding and reconstructed embedding and constraint the property of reconstructed space by introducing the cycle consistency. In this way, the comparison is conducted in the constraint original and reconstructed text spaces. For ReferFormer, it models the alignment of text to video by querying the visual features by text in the transformer decoder. In this way, the comparison is conducted in unconstrained text and video spaces thus results in a inferior performance.
    \item FPS: The speed improvement of our method mainly comes from our efficient multimodal fusion. Compared to the multi-scale CM-FPN, our early-grounding module is only conduct at the high-level. In addition, our bi-direction multimodal fusion (Equ 4 \& 5) only leverages cross-attention to avoid computational expensive video-to-video operations.
\end{itemize}

\section{Visualization of Attentions in OLM}

\begin{figure}[ht]
    \centering
    \includegraphics[width=1\linewidth]{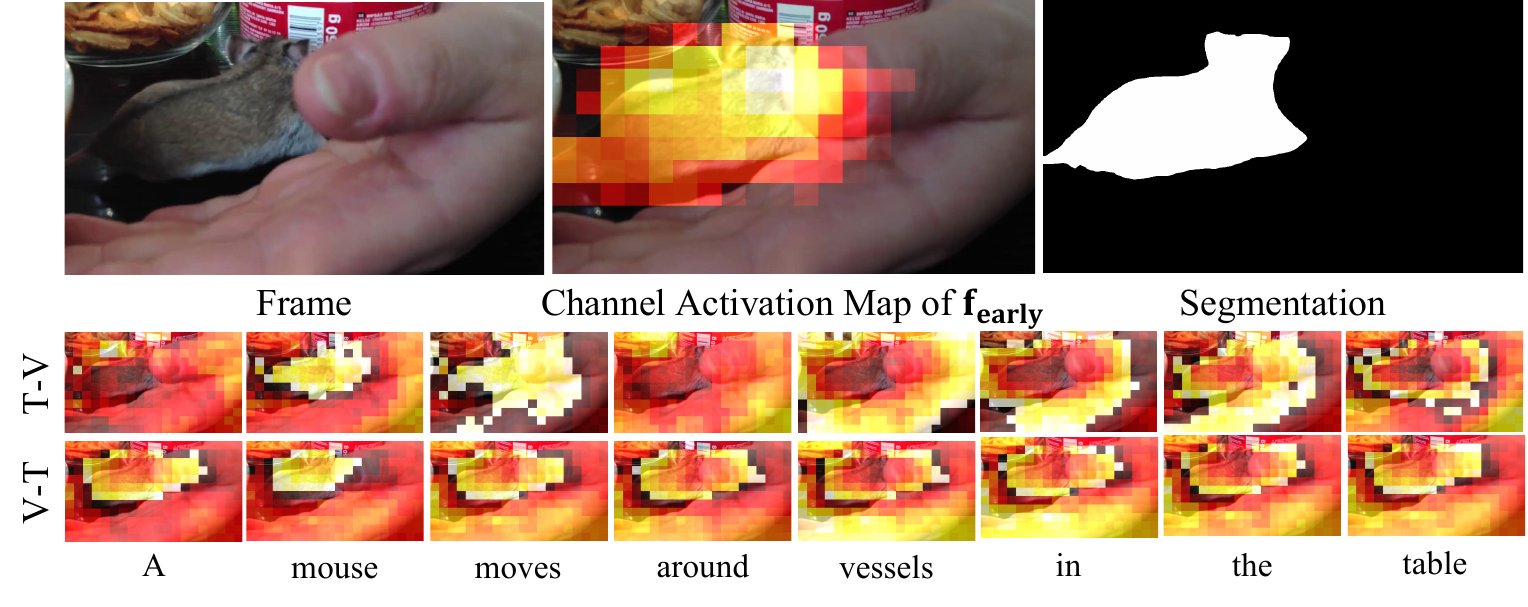}
    \caption{\textbf{Visualization of cross-attention attentions and $\mathrm{\mathbf{f_{early}}}$ in the Early Grounding Module.}}
    \label{fig:decoder}
\end{figure}

\section{More Implementation Details}
We pretrain our model on a combination of three image-level datasets, \ie, Ref-COCO, Ref-COCO+, and Ref-COCOg \cite{yu2016refcoco}. To be compatible with the image-level dataset, we set the window size to 1. We pretrain our model for 12 epochs, which takes about 1-2 days on 8 NVIDIA V100 32G GPUs depending on the backbones. We select the checkpoint with the best results on Ref-COCO val set as our pretrained weight for our main training.

We set the $\lambda_{text}=0.1$, $\lambda_{cls}=2$, $\lambda_{mask}=2$, $\lambda_{align}=1$, $\lambda_{angle}=10$, $\lambda_{L1}=5$, $\lambda_{giou}=2$, $\lambda_{dice}=2$ and $\lambda_{focal}=5$ during all training process. $C_v=C_e=C_q=256$ is utilized. The positional embedding added in the transformers is the standard triangle positional embedding used in \cite{vaswani2017attention}. We set the layer number to three for transformers decoders $\mathcal{D}_e$ and $\mathcal{D}_v$. The dynamic filter number $K$ is set to 3. The data point to calculate the relational loss is selected within each batch for simplicity. The text encoder is frozen during the main training.

\section{Detailed Structure of Mask Decoding}

\begin{figure}[ht]
    \centering
    \includegraphics[width=0.8\linewidth]{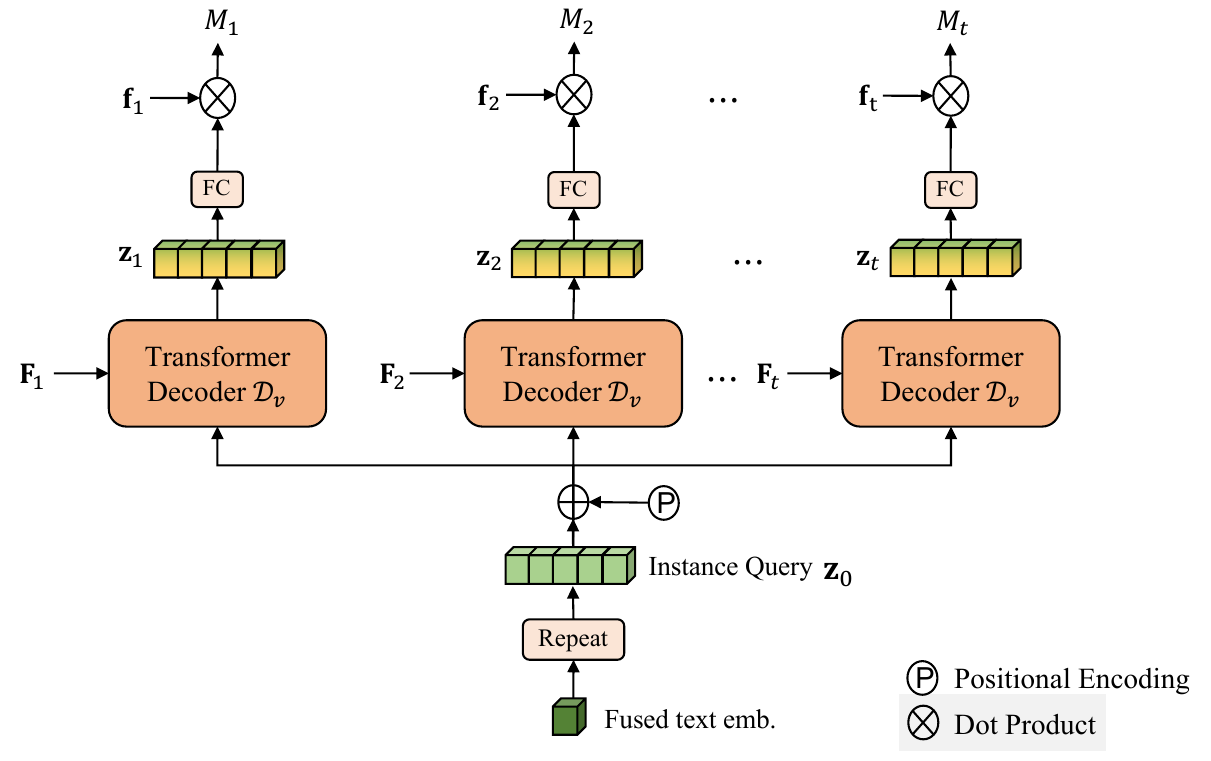}
    \caption{\textbf{Illustration of mask decoding.}}
    \label{fig:decoder}
\end{figure}

As is shown in Fig.~\ref{fig:decoder}, given the fused text embedding, we generate the instance query $\mathrm{\mathbf{z}}_{0}$ by repeating the fused text embedding $N$ times where $N$ is the query number. After that, we generate instance embedding $\{z_t\}_{t=1}^T$ for each time step separately using a shared transformer decoder $\mathcal{D}_{v}$ with encoded memory $\{\mathrm{\mathbf{F}}_t\}_{t=1}^T$ from visual encoder. The mask prediction $M_t$ for each time step $t$ is derived by a linear combination of $\mathrm{\mathbf{F}}_t$ where weights are learned from instance embedding $\mathrm{\mathbf{z}}_t$ by two fully connected layers. Note that, as positional embedding is added to the instance query $z_0\in\mathbb{R}^{C_q\times N}$, each slot in the instance query is different. 


\section{Broader Impact and Future Works}
The false alarm problem in the RVOS task also exists in other referring prediction tasks, \eg, visual grounding \cite{deng2021transvg} and referring image segmentation \cite{ye2019cmsa}. We consider our problem formulation that defines the negative and positive vision-language pairs can be extended to other tasks that require multi-modal semantic consensus.

\bibliographystyle{plain}
\bibliography{src/reference}


%% file: src/0-abstract.tex
\vspace{-0.4cm}
\begin{abstract}
Referring Video Object Segmentation (R-VOS) is a challenging task that aims to segment an object in a video based on a linguistic expression. 
Most existing R-VOS methods have a critical assumption: the object referred to must appear in the video. This assumption, which we refer to as ``\textbf{semantic consensus}", is often violated in real-world scenarios, where the expression may be queried against false videos. In this work, we highlight the need for a robust R-VOS model that can handle semantic mismatches.
Accordingly, we propose an extended task called Robust R-VOS (\rrvos), which accepts unpaired video-text inputs.
We tackle this problem by jointly modeling the primary R-VOS problem and its dual (text reconstruction).
A \textbf{structural} text-to-text cycle constraint is introduced to discriminate semantic consensus between video-text pairs and impose it in positive pairs, thereby achieving multi-modal alignment from both positive and negative pairs.
Our structural constraint effectively addresses the challenge posed by linguistic diversity, overcoming the limitations of previous methods that relied on the point-wise constraint.
A new evaluation dataset, \rrytbvos~is constructed to measure the model robustness.  
Our model achieves state-of-the-art performance on R-VOS benchmarks, Ref-DAVIS17 and Ref-Youtube-VOS, and also our \rrytbvos~dataset.
\end{abstract}

%% file: src/1-introduction.tex
\section{Introduction}

Referring video object segmentation (R-VOS) aims to segment a referred object in a video given a linguistic expression.
R-VOS has witnessed growing interest thanks to its promising potential in human-computer interaction, such as video editing and augmented reality. 


\begin{figure}[t]
    \centering
    \includegraphics[width=\linewidth]{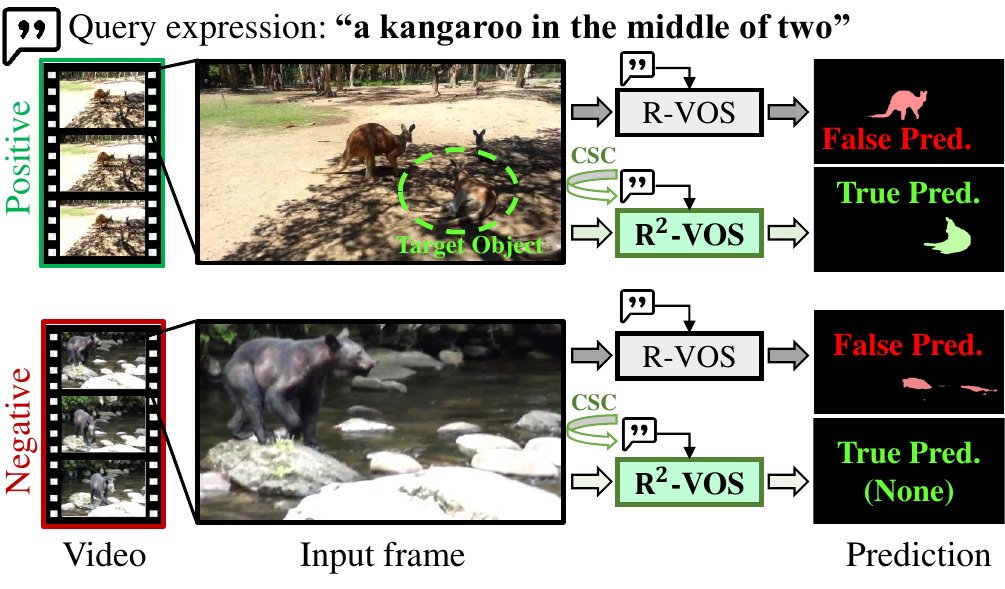}
    \vspace{-0.5cm}
    \caption{\textbf{Illustration of the new \rrvos~task.} A linguistic expression is given to query a set of videos without the semantic consensus assumption in \rrvos. 
   Videos containing the referred object by the expression are \textcolor[rgb]{0,0.6875,0.3125}{positive}, otherwise \textcolor[rgb]{0.6428125,0.047875,0.03225}{negative}. 
   Existing R-VOS methods such as Referformer \cite{referformer} easily segment irrelevant objects in both positive and negative videos, as they do not investigate the consensus. In contrast, our \rrvos~method achieves greater accuracy by incorporating cyclic structural consensus (CSC).
    }
    \vspace{-0.4cm}
    \label{fig:teaser}
\end{figure}



In previous studies \cite{botach2021mttr,referformer,ding2022language,chen2022multi}, the R-VOS problem is addressed with the strict assumption that the referred object appears in the video, thus requiring object-level \textbf{semantic consensus} between the expression and the video. However, this assumption is not always true in practice, leading to severe false-alarm problems in negative videos (referred object does not present in the video), as demonstrated in \cref{fig:teaser}. This limitation hinders the application of such methods in scenarios where matched vision-language pairs are not available. We argue that the current R-VOS task is incomplete when assuming that the referred object always exists in the video.


Even with semantic consensus in positive video-language pairs, locating the correct object is still challenging due to its multimodal nature. Recent method MTTR \cite{botach2021mttr} employs multimodal transformer encoders to learn a joint representation, and localizes the object by ranking all presented objects in the video. ReferFormer \cite{referformer} and the image method ReTR \cite{li2021referring} adopt the linguistic expression as a query to the transformer decoder to avoid redundant ranking of all objects. However, these latest R-VOS methods still suffer from a semantic misalignment of the segmented object and the linguistic expression, as shown in \cref{fig:teaser}. Some previous works \cite{mao2016generation,chen2018knowledge,chen2019referring} attempt to enhance the alignment by imposing cycle consistency between referring expressions and their reconstructed counterparts, without considering negative vision-language pairs. Moreover, they restrict the two expressions to be identical or similar, 
which may not always be the case due to linguistic diversity. Expressions that describe an object can vary from different perspectives, \eg, the target kangaroo in \cref{fig:teaser} can be described either as `` in the middle of two'' or `` lying on its side''.
Therefore, a more robust cycle constraint is required to better accommodate this diversity.

In this paper, we seek to investigate the semantic alignment problem between visual and linguistic modalities in referring video segmentation.
We extend the current R-VOS task to accept unpaired video-language inputs.
This new task, termed \textbf{Robust R-VOS (\rrvos)}, goes beyond the limitation of the existing R-VOS task by analyzing the semantic consensus. It additionally discriminates between \textbf{positive} video-language pairs where the referred object appears in the video, and \textbf{negative} pairs where it does not.

The \rrvos~task essentially corresponds to two interrelated problems: the \textbf{primary} problem of segmenting masks in videos with referring texts (R-VOS), and its \textbf{dual} problem of reconstructing text expressions from videos with object masks. To jointly optimize both problems, consistency in the text-to-text cycle is usually imposed if the video-language pairs are positive, as in most previous works~\cite{mao2016generation,chen2018knowledge,chen2019referring}. We go further by learning from both positive and negative pairs. Specifically, we investigate the semantic consensus using cycle consistency measure to discriminate the positive and negative pairs, while also imposing consistency in positive pairs. Both the discrimination and imposing procedure help with the video-language alignment learning. Furthermore, instead of using point-wise cycle consistency for individual samples~\cite{chen2018knowledge,chen2019referring}, we adopt a \textbf{structural} measurement to assess the relation consistency between the referring textual space and its reconstructed counterpart.
This design accommodates linguistic diversity better, as it considers that an object may have different text expressions but should be differentiated from the expressions describing other objects. 
In addition, we enable the end-to-end joint training of the primary and dual problems by introducing an object localizing module (OLM) as a proxy to bridge the two problems. 
The dual text reconstruction can be conducted using the proxy features without waiting for the resulting masks from the primary task.


Our contributions are summarized as: 
\begin{itemize}
\setlength{\itemsep}{1pt}
\setlength{\parsep}{0pt}
\setlength{\parskip}{0pt}
    \item We are the first to address the severe false-alarm problem faced by previous R-VOS methods with unpaired video-language inputs. 
    We introduce the new \rrvos~task accepting unpaired inputs, as well as an evaluation dataset and corresponding metrics.
    \item We introduce the cyclic structural consensus for both discrimination between positive and negative video-language pairs and enhancement of segmentation quality, which better accommodates linguistic diversity.
    \item We propose a \rrvos~network that enables end-to-end training of the primary referring segmentation and dual expression reconstruction task by introducing the object localizing module as a proxy.
    \item Our method achieves state-of-the-art performance for both R-VOS and $\mathrm{R}^2$-VOS tasks on popular Ref-Youtube-VOS, Ref-DAVIS, and new $\mathrm{R}^2$-Youtube-VOS datasets.
\end{itemize}


%% file: src/2-related_works.tex
\section{Related Works}

\begin{figure*}[t]
    \centering
    \vspace{-0.2cm}
    \includegraphics[width=1\linewidth]{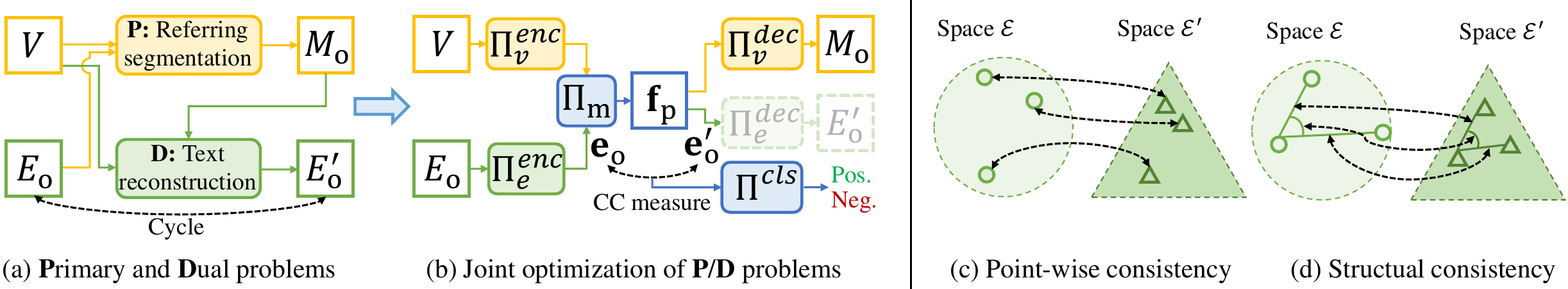}
    \vspace{-0.5cm}
    \caption{\textbf{Problem analysis.}
    (a) \rrvos~explores the \textbf{P}rimary problem of referring segmentation and the \textbf{D}ual problem of text reconstruction.
    The \textbf{P}/\textbf{D} problems are linked in a cycle path from the input expression $E_o$ to the reconstructed expression $E'_o$.
    (b) We utilize the cycle consistency (CC) measure for 1) discriminating positive/negative pairs with module $\Pi^{cls}$, 2) alignment enhancement for positive pairs. We enable the joint optimization of \textbf{P}/\textbf{D} problems with a cross-modal proxy $\mathrm{\mathbf{f}}_{p}$ between all uni-modal operations (\ie,  $\Pi_v^{enc}$, $\Pi_e^{enc}$, $\Pi_v^{dec}$ and $\Pi_e^{dec}$).
    (c) Point-wise consistency is not suitable in \rrvos~as the mapping between $\mathcal{E}$ and $\mathcal{E}'$ are not necessarily bijective (an object can be described with various expressions).
    (d) Instead, we apply a structural consistency to preserve relationships (distances and angles).
     }
    \vspace{-0.2cm}
    \label{fig:rrvos}
\end{figure*}

\vspace{0.2cm}\noindent\textbf{Referring segmentation.}\,
Referring image segmentation (RIS) and R-VOS aim to segment an object in an image or video sequence respectively, given a linguistic description as the query. Recent RIS methods \cite{LAVT,VLT} achieves promising results by using the power of multimodal transformer.
Going beyond the RIS task \cite{hu2016segmentation,li2018referring, yu2018mattnet, CMPC, MCN, BUSNet, CGAN}, R-VOS is a more challenging task as it requires leveraging both intra-frame and temporal cues. URVOS \cite{seo2020urvos} is the first unified R-VOS framework with a cross-modal attention and a memory attention module, which largely improves R-VOS performance. 
ClawCraneNet \cite{liang2021clawcranenet} leverages cross-modal attention to bridge the semantic correlation between textual and visual modalities. 
ReferFormer \cite{referformer} and MTTR \cite{botach2021mttr} are two latest works that utilize transformers to decode or fuse multimodal features. ReferFormer \cite{referformer} employs a linguistic prior to the transformer decoder to focus on the referred object. MTTR \cite{botach2021mttr} leverages a multimodal transformer encoder to fuse linguistic and visual features. Different from other vision-language tasks, \eg, image-text retrieval \cite{lin2014microsoft,liu2019use,miech2018learning} and video question answering \cite{lei2018tvqa,song2018explore}, R-VOS needs to construct object-level multimodal semantic consensus in a dense visual representation.


\vspace{0.2cm}\noindent\textbf{Consistency constraint in multi-modal learning.}\,
The pioneer work \cite{mao2016generation} jointly model generation and comprehension of linguistic object description in images and find it benefits both tasks.
KACNet \cite{chen2018knowledge} improves visual grounding by enhancing semantic alignment with knowledge-aided consistency constraints. 
\cite{chen2019referring} also utilize caption-aware consistency to improve referring image segmentation.
All these methods use text-to-text (point-wise) consistency, assuming that the referring text and reconstructed text are similar. However, in the real world, an object can be described with diverse expressions from different perspectives. In our work, we do not enforce the cycle consistency between two expressions but rather the structural consistency between two expression spaces.
Moreover, all these methods only utilize cycle consistency to enhance aligning multi-modal features. They tend to fail when the referred object is not presented in the image, as discussed in \cite{mao2016generation}.


%% file: src/3-problem.tex
\section{Robust R-VOS}
\subsection{Problem Definition}
We introduce a novel task, robust referring video segmentation (\rrvos), which aims to predict mask sequences $\{M_o\}$ for an unconstrained video set $\{V\}$ given an expression $E_o$ of an object $o$. Different from the setting of the previous R-VOS problem, \rrvos~does not impose the constraint that the queried videos must contain the object specified by the expression $E_o$.
A video $V$ and an expression $E_o$ have \textbf{semantic consensus} if the object $o$ referred to by $E_o$ appears in $V$. Then, the video is considered \textit{positive} with respect to $E_o$, otherwise, it is \textit{negative}.
In the \rrvos task, the goal is to discriminate between positive and negative videos, and predict masks $M_o$ of the target object $o$ for positive videos. For negative videos, all frames are treated as background.

\noindent\textbf{Primary problem.}
The referring segmentation can be formulated as the maximum {\em a posteriori} estimation problem of $p(M_o|V, E_o)$. By applying the Bayes rule, we obtain:
\begin{equation}
    p(M_o|V,E_o) \sim p(E_o|V,M_o)p(M_o|V)
\end{equation}
\noindent\textbf{Dual problem.}
As the prior $p(M_o|V)$ is not affected by the expression $E_o$, we consider maximizing $p(E_o|V, M_o)$ as a dual problem of the referring segmentation (primary problem), which is to reconstruct the text expression given the video and object masks. 

\noindent\textbf{Joint optimization.}
The primary and dual problem can be linked in a cyclic path, as previous works \cite{shi2020query,chen2019referring}, between the input expression $E_o$ and the reconstructed expression $E'_o$, as depicted in \cref{fig:rrvos} (a). Previous methods only consider positive text-video pairs and impose consistency between $E_o$ and $E'_o$ for all samples without differentiating between positive and negative pairs.
To go a step further, we consider that the negative pairs can also contribute to joint optimization. We model the semantic consensus for each text-video pair with a consistency measure that allows the text-video alignment to benefit from imposing cycle consistency in positive pairs and also from discrimination between positive and negative pairs.


To avoid the impact of pretrained language models and enable end-to-end training, we investigate the consistency measurement between the original textual embedding space $\mathcal{E}$ and the reconstructed textual embedding space $\mathcal{E}'$. 
To link the primary and dual problem and facilitate the joint optimization, we introduce a cross-modal proxy feature $\mathrm{\mathbf{f}}_{p}$ that captures information from both the inputs of the primary and dual problems, \ie, $(V, E_o)$ and $(V, M_o)$, after multi-modal interaction $\Pi_m$, as shown in \cref{fig:rrvos} (b).  $\mathrm{\mathbf{f}}_p$ is placed between the unimodal encoders and decoders, \ie, $\Pi_v^{enc}$, $\Pi_e^{enc}$, $\Pi_v^{dec}$, $\Pi_e^{dec}$, to focus solely on the multimodal interaction. An additional module $\Pi^{cls}$ for discriminating between positive and negative pairs.

\subsection{Cyclic Consensus Measurement}
As mentioned, we utilize the text-to-text cycle consistency to measure the text-video semantic consensus.

\noindent\textbf{Point-wise cycle consistency.} Previous works compute point-wise cycle consistency for each individual sample (\cref{fig:rrvos} (c)), \ie, $CC(\mathbf{e}_i, \mathbf{e}'_i), \forall \mathbf{e}_i \in \mathcal{E}, \mathbf{e}'_i \in \mathcal{E}'$, where $CC$ denotes a distance measure function. 
However, we observe that the mapping between the two spaces $\mathcal{E}$ and $\mathcal{E}'$ is not necessarily bijective, as there could be multiple textual descriptions for the same object from different perspectives. As a result, $\mathbf{e}_i$ and $\mathbf{e}'_i$ may not be close. Thus, simply utilizing point-wise consistency will collapse the feature space to be sub-optimal.

\begin{figure*}[t]
    \centering
    \includegraphics[width=\linewidth]{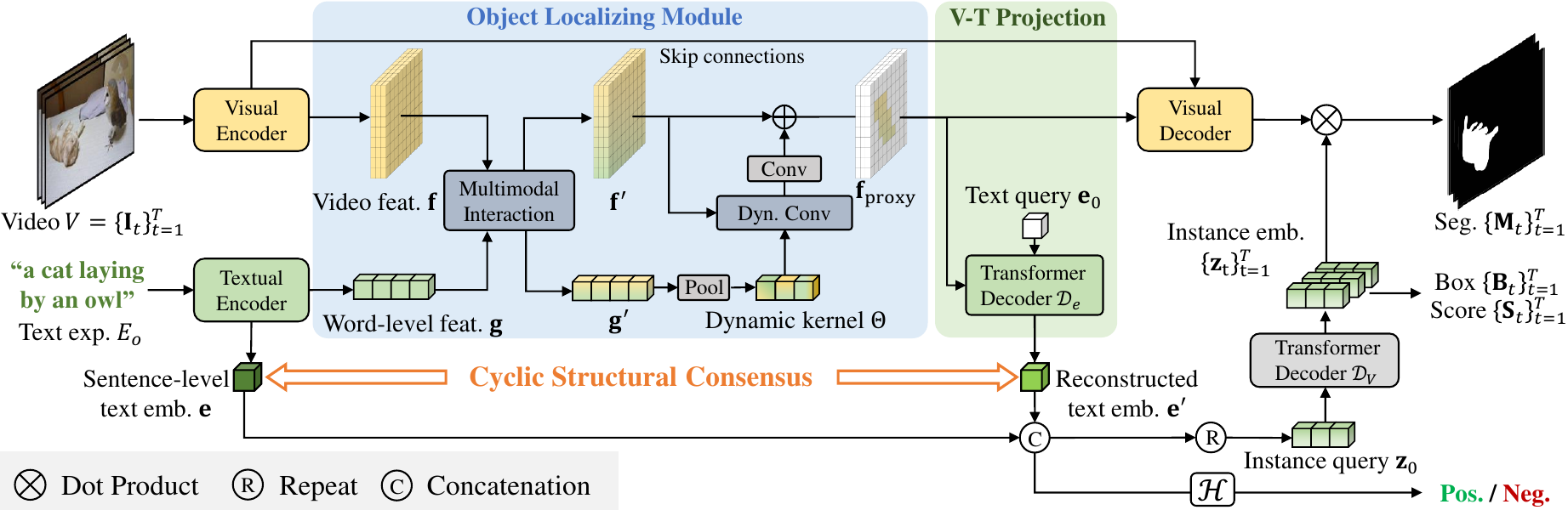}
    \vspace{-0.3cm}
    \caption{\textbf{Overview of the \rrvos~network}. Given a video clip $V=\{\mathbf{I}_t\}_{t=1}^T$ and a textual expression $E_o$ referring to object $o$, we first extract unimodal features $\mathrm{\mathbf{f}}$ and $\mathrm{\mathbf{g}}$, then fuse them in the object localizing module to obtain the proxy feature $\mathrm{\mathbf{f_{proxy}}}$. Then, $\mathrm{\mathbf{f_{proxy}}}$ is projected to a textual space as $\mathrm{\mathbf{e}}^\prime$. The semantic consensus assessed with cyclic structural consistency between $\mathrm{\mathbf{e}}^\prime$ and $\mathrm{\mathbf{e}}$ is used for discriminating positive/negative pairs with $\mathcal{H}$, and imposed for positive pairs. The final segmentation is obtained by dynamic convolutions with features from the visual decoder and dynamic weights from $\mathrm{\mathbf{e}}^\prime$ and $\mathrm{\mathbf{e}}$. 
    }
    \vspace{-0.4cm}
    \label{fig:pipeline}
\end{figure*}

\noindent\textbf{Structural cycle consistency.}
In contrast, we consider that the structure of the reconstructed embedding space $\mathcal{E}'$ should be preserved as that of $\mathcal{E}$. Although an object may have different text expressions, they should be differentiated from the expressions describing other objects. Thus, the relationship between expressions that describe different objects should be preserved, as illustrated in \cref{fig:rrvos} (d).
Mathematically, the structure-preserving property is defined as isometric and conformal constraints to preserve pair-wise distances and angles for $\mathbf{e} \in \mathcal{E}$ and $\mathbf{e}' \in \mathcal{E}'$:
\begin{eqnarray}
\label{eq:relation_cycle_consistency}
    & CC_{dist}(|\mathbf{e}_1 - \mathbf{e}_2|, |\mathbf{e}'_1 - \mathbf{e}'_2|), \\
    & CC_{angle} (\angle (\overrightarrow{\mathbf{e}_2\mathbf{e}_1}, \overrightarrow{\mathbf{e}_2\mathbf{e}_3}) , \angle (\overrightarrow{\mathbf{e}'_2\mathbf{e}'_1}, \overrightarrow{\mathbf{e}'_2\mathbf{e}'_3})),
\end{eqnarray}
where $|\cdot|$ denotes a distance metric,  $CC_{dist}$ and $CC_{angle}$ denote cycle consistency measurement function for distances and angles, respectively. The detailed implementation is elaborated in the loss function (\cref{sec:loss}).





%% file: src/4-method.tex
\section{R\textsuperscript{2}-VOS Network}
In this section, we elaborate on the structure of \rrvos~network, which mainly consists of four parts: feature extraction, object localizing module (OLM), video-text (V-T) projection for text reconstruction, and mask decoding for final segmentation, as illustrated in \cref{fig:pipeline}.
We first extract the video feature $\mathrm{\mathbf{f}}$, word-level text feature $\mathrm{\mathbf{g}}$, and sentence-level text feature $\mathrm{\mathbf{e}}$.
On the one hand, to model the primary segmentation problem of maximizing $p(M_o|V, E_o)$, 
we enable the multimodal interaction in OLM to generate the grounded feature $\mathrm{\mathbf{f_{proxy}}}$.
$\mathrm{\mathbf{f_{proxy}}}$ coarsely locates the referred object $o$ with irrelevant features suppressed, which serves as a proxy linking the primary segmentation and dual text reconstruction problem.
The final mask $M_o$ is obtained by dynamic convolution \cite{chen2020dynamic} on the decoded visual feature maps, with kernels learned from instance embedding $\{\mathrm{\mathbf{z}_t}\}_{t=1}^{T}$.
On the other hand, to model the dual text reconstruction problem of maximizing $p(E_o|V,M_o)$, we utilize $\mathrm{\mathbf{f_{proxy}}}$ as the alternative of $V$ and $M_o$, since $\mathrm{\mathbf{f_{proxy}}}$ conveys grounded video clues of object $o$.
In this way, we enable the parallel optimization of the primary and dual problem by relating them to $\mathrm{\mathbf{f_{proxy}}}$.
Specifically, we employ a V-T projection module to project $\mathrm{\mathbf{f_{proxy}}}$ onto a reconstructed text embedding $\mathrm{\mathbf{e}}^\prime$.
We add a cyclic structural constraint between $\mathrm{\mathbf{e}}^\prime$ and  $\mathrm{\mathbf{e}}$ to enforce the semantic alignment between the segmented mask and expression for positive videos. 
In addition, we introduce a classification head $\mathcal{H}(\mathrm{\mathbf{e}},\mathrm{\mathbf{e}}^\prime)$, which discriminates the semantic consensus by assessing the consistency between $\mathrm{\mathbf{e}}$ and $\mathrm{\mathbf{e}}^\prime$.

\subsection{Unimodal Feature Extraction}
\vspace{0.2cm}\noindent\textbf{Visual encoder.}
We build the visual encoder with a visual backbone and a deformable transformer encoder \cite{zhu2020deformable} on top of it, as that in \cite{referformer}. 
The extracted features from the backbone are flattened, projected to a lower dimension, added with positional encoding \cite{ke2020rethinking}, and then fed into a deformable transformer encoder \cite{zhu2020deformable} similar to the previous method~\cite{referformer}. We denote the multi-scale output of the transformer encoder as $\mathrm{\mathbf{F}}$ and the low-resolution visual feature map from the backbone as $\mathrm{\mathbf{f}}$, where $\mathrm{\mathbf{f}} \in \mathbb{R}^{T \times C_{v} \times \frac{H}{32} \times \frac{W}{32}}$, $C_v$ is the feature channel, $T$ is the video length and $H$ and $W$ are the original image size.

\vspace{0.2cm}\noindent\textbf{Textual encoder.} We leverage a pre-trained linguistic model RoBERTa \cite{liu2019roberta} to map the input textual expression $E_o$ to a textual embedding space. The textual encoder extracts a sequence of word-level text feature $\mathrm{\mathbf{g}}\in\mathbb{R}^{C_{e}\times L}$ and a sentence-level text embedding $\mathrm{\mathbf{e}}\in\mathbb{R}^{C_{e}\times 1}$, where $C_{e}$ and $L$ are the dimension of linguistic embedding space and the expression length respectively.

\subsection{Object Localizing Module}
\begin{figure}[t]
    \centering
    \vspace{-0.2cm}
    \includegraphics[width=\linewidth]{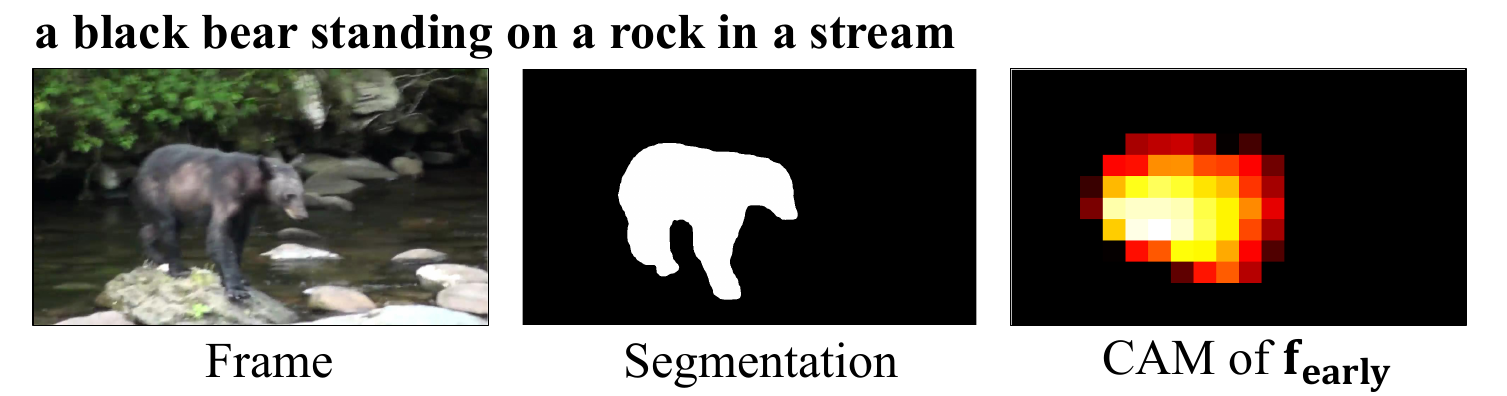}
    \vspace{-0.5cm}
    \caption{\textbf{Channel activation map} (CAM) of $\mathrm{\mathbf{f_{proxy}}}$.}
    \vspace{-0.2cm}
    \label{fig:vis_cam}
\end{figure}

We employ the object localizing module (OLM) to coarsely locate the referred object $o$.
The grounded feature $\mathbf{f_{proxy}}$ encoding information of $o$ can not only be utilized for the primary segmentation problem, but also for the dual expression reconstruction task, which serves as a proxy connecting the two problems. 
Specifically, we utilize dynamic convolution~\cite{chen2020dynamic} to discriminate visual features in the early stage.
As shown in the blue part of \cref{fig:pipeline}, we first enable the multimodal interaction between video and text features, then apply the dynamic convolution with kernels learned from text features to discriminate the object-level semantics.
In particular, multi-head cross-attention (MCA) \cite{vaswani2017attention} is leveraged to conduct the multimodal interaction:
\begin{equation}
\begin{aligned}
    &\mathrm{\mathbf{h_f}}=\mathrm{LN}( \mathrm{MCA}(\mathrm{\mathbf{f}}, \mathrm{\mathbf{g}}) + \mathrm{\mathbf{f}}) \quad \mathrm{\mathbf{f^\prime}}=\mathrm{LN}(\mathrm{FFN}(\mathrm{\mathbf{h_f}}) + \mathrm{\mathbf{h_f}}) \\
    &\mathrm{\mathbf{h_g}}=\mathrm{LN}( \mathrm{MCA}(\mathrm{\mathbf{g}}, \mathrm{\mathbf{f}}) + \mathrm{\mathbf{g}}) \quad
    \mathrm{\mathbf{g^\prime}}=\mathrm{LN}(\mathrm{FFN}(\mathrm{\mathbf{h_g}}) + \mathrm{\mathbf{h_g}})
\end{aligned}
\end{equation}
where $\mathrm{MCA}(\mathrm{\mathbf{X,Y}})$ denotes $\mathrm Attention(\mathrm{\mathbf{W^Q}}\mathrm{\mathbf{X}},\mathrm{\mathbf{W^K}}\mathrm{\mathbf{Y}},$ $\mathrm{\mathbf{W^V}}\mathrm{\mathbf{Y}})$. $\mathrm{\mathbf{W}}$ represents learnable weight. $\mathrm{LN}$ and $\mathrm{FFN}$ denote layer normalization and feed-forward network respectively.
The word-level text feature $\mathrm{\mathbf{g^\prime}}$ is further pooled to a fixed length, and followed by a fully-connected layer to form the word kernels $\Theta=\{\theta_i\}_{i=1}^{K}$. $K$ is the kernel number and $\theta_i\in\mathbb{R}^{C\times 1}$. 
The dynamic kernels are applied separately to video feature $\mathrm{\mathbf{f^\prime}}\in\mathbb{R}^{C\times T HW}$ to form the $\mathrm{\mathbf{f_{proxy}}}\in\mathbb{R}^{C\times T HW}$: 
\begin{equation}
    \mathrm{\mathbf{f_{proxy}}} = \mathrm{BN}(\varphi(\theta_1\mathrm{\mathbf{f^\prime}}\oplus\dots\oplus\theta_K\mathrm{\mathbf{f^\prime}}) + \mathrm{\mathbf{f^\prime}}),
\end{equation}
where $\oplus$ is the concatenation in channel dimension and $\varphi(\cdot)$ is a $1\times1$ convolution to reduce the dimension. $\mathrm{BN}$ denotes batch normalization. \cref{fig:vis_cam} shows a visualization of $\mathrm{f_{proxy}}$. The channel activation map of $\mathrm{f_{proxy}}$ indicates that the OLM can effectively locate the referred object and suppress irrelevant features.

\subsection{Text Reconstruction}
\vspace{0.2cm}\noindent\textbf{V-T projection.}
We leverage a transformer decoder $\mathcal{D}_E$ as the textual decoder to transform the visual representation of the referred object into the textual space. As shown in \cref{fig:pipeline}, a learnable text query $\mathrm{\mathbf{e}}_0\in\mathbb{R}^{C_e\times 1}$ is employed to query the $\mathrm{\mathbf{f_{proxy}}}$. The output of the transformer decoder is the reconstructed text embedding $\mathrm{\mathbf{e}}^\prime=\mathcal{D}_E(\mathrm{\mathbf{f_{proxy}}},\mathrm{\mathbf{e}}_0)\in\mathbb{R}^{C_e\times 1}$.

\subsection{Referring Segmentation}

\vspace{0.2cm}\noindent\textbf{Mask segmentation.}
We leverage deformable transformer decoders with dynamic convolution to segment the object masks as that in \cite{referformer}.
Since the reconstructed text embedding is generated with visual features injected, we consider it can encode some visual information, thus augmenting the original text embedding. As shown in \cref{fig:pipeline}, we first fuse the reconstructed text embedding $\mathrm{\mathbf{e}}^\prime$ to text embedding $\mathrm{\mathbf{e}}$. The fused text embedding $\mathrm{\mathbf{e}}$ is then repeated $N$ times to form the instance query \cite{wang2021end} $\mathrm{\mathbf{z}}_0\in\mathbb{R}^{C_q\times N}$, where $C_q$ is the dimension of instance query and $N$ is the instance query number.
We then use $T\times$ deformable transformer decoders $\mathcal{D}_V$ with shared weights to decode the instance embeddings $\mathrm{\mathbf{z}}_t\in\mathbb{R}^{C_q\times N}$ for each frame, \ie, $\mathrm{\mathbf{z}}_t=\mathcal{D}_V(\mathrm{\mathbf{F}}_t,\mathrm{\mathbf{z}}_0)$. $\mathrm{\mathbf{F}}_t$ is the multiscale visual feature from visual encoder at time $t$. 
A dynamic kernel $\mathrm{\mathbf{w}}_t$ is further learned from the instance embedding $\mathrm{\mathbf{z}}_t$.
The final feature map $\mathrm{\mathbf{f_{out}}}_{,t}\in\mathbb{R}^{C\times H\times W}$ is obtained by fusing low-level features from the feature pyramid network \cite{lin2017feature} in the visual decoder.
The mask prediction $\mathrm{\mathbf{M}}_t\in\mathbb{R}^{N\times H\times W}$ can be computed by $\mathrm{\mathbf{M}}_t =\mathrm{\mathbf{w}}_t^\mathrm{T}\mathrm{\mathbf{f_{out}}_{,t}}$.

\vspace{0.2cm}\noindent\textbf{Auxiliary heads.}
We build a set of auxiliary heads to obtain the final object masks across frames. In particular, a box head, a scoring head and a semantic consensus discrimination head are leveraged to predict the bounding boxes $\mathrm{\mathbf{B}}_t\in\mathbb{R}^{N\times 4}$, confidence scores $\mathrm{\mathbf{S}}_t\in\mathbb{R}^{N\times 1}$ and the alignment degree of multimodal semantics $A\in\mathbb{R}$. The box and scoring head are two fully-connected layers upon the instance embedding $\mathrm{\mathbf{e_t}}$. 
The semantic consensus discrimination head $\mathcal{H}(\mathrm{\mathbf{e}},\mathrm{\mathbf{e}}^\prime)$ consists of two fully-connected layers upon the text embeddings $\mathrm{\mathbf{e}}\oplus \mathrm{\mathbf{e}}^\prime$. 
Note that $A$ represents the semantic alignment in the entire video rather a single frame, since the expression is a video-level description. 


\subsection{Loss Function}
\label{sec:loss}
The loss function of our method can be boiled down to three parts:
\begin{equation}
    \mathcal{L} = \lambda_{text}\mathcal{L}_{text} + \lambda_{segm}\mathcal{L}_{segm} + \lambda_{align}\mathcal{L}_{align},
\end{equation}
where $\mathcal{L}_{text}$, $\mathcal{L}_{segm}$, and $\mathcal{L}_{align}$ are losses for text reconstruction, referring segmentation and semantic consensus discrimination respectively. 

\vspace{0.2cm}\noindent\textbf{Loss of consensus discrimination.}
$\mathcal{L}_{align}$ is a cross-entropy loss between the predicted alignment $A$ and ground-truth $\hat{A}$. $\hat{A}=1$ represents positive video-language pairs, otherwise 0. 

\vspace{0.2cm}\noindent\textbf{Loss for text reconstruction.}
Given the text embedding $\mathrm{\mathbf{e}}$ and reconstructed embedding $\mathrm{\mathbf{e}}^\prime$, we use a structural constraint to impose the cycle consistency between $\mathrm{\mathbf{e}}$ and $\mathrm{\mathbf{e}}^\prime$ for positive pairs. We calculate the loss by 
\begin{equation}
\mathcal{L}_{text} =\hat{A}\cdot(\mathcal{L}_{dist} + \lambda_{angle}\mathcal{L}_{angle}), 
\end{equation}
where $\lambda_{angle}$ is a hyperparameter balancing the distance loss $\mathcal{L}_{dist}$ and angle loss $\mathcal{L}_{angle}$.
We elaborate these two losses according to the structural cycle consistency n in \cref{eq:relation_cycle_consistency}.
Let $\mathcal{X}^n = \{ (x_1, ..., x_n)| x_i \in \mathcal{X} \}$ denote a set of $n$-tuples, $\Phi^n = \{(\mathbf{x}, \mathbf{x}') | \mathbf{x} \in \mathcal{X}^n, \mathbf{x}' \in \mathcal{X}'^n \}$ denote a set of pairs consisting of two $n$-tuples of distinct elements from two different sets $\mathcal{X}$ and $\mathcal{X}'$.
Specifically, the distance-based and angle-based relations relate text embeddings of 2-tuple and 3-tuple respectively, \ie,
$\Phi^2=\{(\mathrm{\mathbf{x}},\mathrm{\mathbf{x}}^\prime)|\mathrm{\mathbf{x}}=(\mathrm{\mathbf{e}}_i,\mathrm{\mathbf{e}}_j),\mathrm{\mathbf{x}}^\prime=(\mathrm{\mathbf{e}}^\prime_i,\mathrm{\mathbf{e}}^\prime_j),i\neq j\}$ and $\Phi^3=\{(\mathrm{\mathbf{x}},\mathrm{\mathbf{x}}^\prime)|\mathrm{\mathbf{x}}=(\mathrm{\mathbf{e}}_i,\mathrm{\mathbf{e}}_j,\mathrm{\mathbf{e}}_k),\mathrm{\mathbf{x}}^\prime=(\mathrm{\mathbf{e}}^\prime_i,\mathrm{\mathbf{e}}^\prime_j,\mathrm{\mathbf{e}}^\prime_k),i\neq j\neq k\}$. Then the losses are given by:
\begin{equation}
      \mathcal{L}_{dist} = \sum_{\small{(\mathbf{x},\,\mathbf{x}^\prime)\in\Phi^2}}l_\delta(\phi_D(\mathrm{\mathbf{x}^\prime})),\phi_D(\mathrm{\mathbf{x}^\prime}))
\end{equation}
\begin{equation}
    \mathcal{L}_{angle} = \sum_{(\mathbf{{x}},\, \mathbf{{x}}^\prime)\in \Phi^3}l_\delta(\phi_{\angle}(\mathrm{\mathbf{x}}),\phi_{\angle}(\mathrm{\mathbf{x}}^\prime))
\end{equation}
where $\phi_D(\mathrm{\mathbf{x}})=\frac{1}{\mu (\mathbf{x})}\|\mathrm{\mathbf{e}}_i-\mathrm{\mathbf{e}}_j\|_2$, $\phi_{\angle}(\mathrm{\mathbf{x}})=\mathrm{cos}\angle(\mathrm{\mathbf{e}}_i, \mathrm{\mathbf{e}}_j, \mathrm{\mathbf{e}}_k)$ and $\mu(\mathbf{x})=\sum_{\mathbf{x}=(x_1, x_2) \in \mathcal{X}^2} \frac{|| x_1 - x_2||_2}{ |\mathcal{X}^2|}$ is the average distance function. The Huber loss $l_\delta(x, x') = \frac{1}{2}(x-x')^2$
if $|x-x'|\leq 1$, otherwise $|x-x'|-\frac{1}{2}$. 

\vspace{0.2cm}\noindent\textbf{Loss for referring segmentation.}
Given a set of predictions $\mathrm{\mathbf{y}}=\{\mathrm{\mathbf{y}}_i\}_{i=1}^N$ and ground-truth $\mathrm{\mathbf{\hat{y}}}$, where $\mathrm{\mathbf{y}}_i=\{\mathrm{\mathbf{B}}_{i,t}, \mathrm{\mathbf{S}}_{i,t}, \mathrm{\mathbf{M}}_{i,t}\}_{t=1}^{T}$ and $\mathrm{\mathbf{\hat{y}}}=\{\mathrm{\mathbf{\hat{B}}}_{t}, \mathrm{\mathbf{\hat{S}}}_{t}, \mathrm{\mathbf{\hat{M}}}_{t}\}_{t=1}^{T}$, we search for an assignment $\sigma \in\mathcal{P}_N$ with the highest similarity where $\mathcal{P}_N$ is a set of permutations of N elements ($\mathrm{\mathbf{\hat{y}}}$ is padded with $\emptyset$). The similarity can be computed as 
\begin{equation}
    \begin{aligned}
    \mathcal{L}_{match}(\mathrm{\mathbf{y}}_i,\mathrm{\mathbf{\hat{y}}}) = \lambda_{box}\mathcal{L}_{box}+\lambda_{conf}\mathcal{L}_{conf}
    \\
    +\lambda_{mask}\mathcal{L}_{mask},
    \end{aligned}
\end{equation}
where $\lambda_{box}$, $\lambda_{conf}$, and $\lambda_{mask}$ are weights to balance losses. Following previous works \cite{pminet,wang2021end}, we leverage a combination of Dice \cite{li2019dice} and BCE loss as $\mathcal{L}_{mask}$, focal loss \cite{lin2017focal} as $\mathcal{L}_{conf}$, and GIoU \cite{rezatofighi2019giou} and L1 loss as $\mathcal{L}_{box}$. The best assignment $\hat{\sigma}$ is solved by the Hungarian algorithm \cite{kuhn1955hungarian}. Given the best assignment $\hat{\sigma}$, the segmentation loss between ground-truth and predictions is defined as $\mathcal{L}_{segm}=\mathbbm{1}(\hat{A})\cdot\mathcal{L}_{match}(\mathrm{\mathbf{y}}, \mathrm{\mathbf{\hat{y}}}_{\hat{\sigma}(i)})$.

\subsection{Inference}
During inference, we select the candidate with the highest confidence to predict the final masks:
\begin{equation}
    \begin{aligned}
    \{\mathrm{\mathbf{\bar{M}}}_t\}_{t=1}^T = \{\mathbbm{1}(A>0.5)\cdot\mathrm{\mathbf{M}}_{\bar{s},t}\}_{t=1}^T \\ \mathrm{\mathbf{\bar{s}}}=\argmax\limits_{i}\{\mathrm{\mathbf{S}}_{i,1}+\dots+ \mathrm{\mathbf{S}}_{i,T}\}_{i=1}^N,
    \end{aligned}
    \vspace{-0.1cm}
\end{equation}
where $\{\mathrm{\mathbf{\bar{M}}}_t\}_{t=1}^T$ is the masks of referred object. $\mathrm{\mathbf{S}}_{i,t}$ and $\mathrm{\mathbf{M}}_{i,t}$ represent the $i$-th candidate in $\mathrm{\mathbf{S}}_t$ and $\mathrm{\mathbf{M}}_{t}$. $\mathrm{\mathbf{\bar{s}}}$ is the slot with the highest confidence to be the target object. We use $\mathbbm{1}(A)$ to filter out predictions in negative videos to mitigate false alarms. $\mathbbm{1}(A>0.5)=1$ if $A>0.5$, else 0.

%% file: src/5-experiments.tex
\vspace{-0.1cm}
\section{Experiment}

\begin{figure*}[t]
    \centering
    \vspace{-0.2cm}
    \includegraphics[width=\linewidth,height=0.32\linewidth]{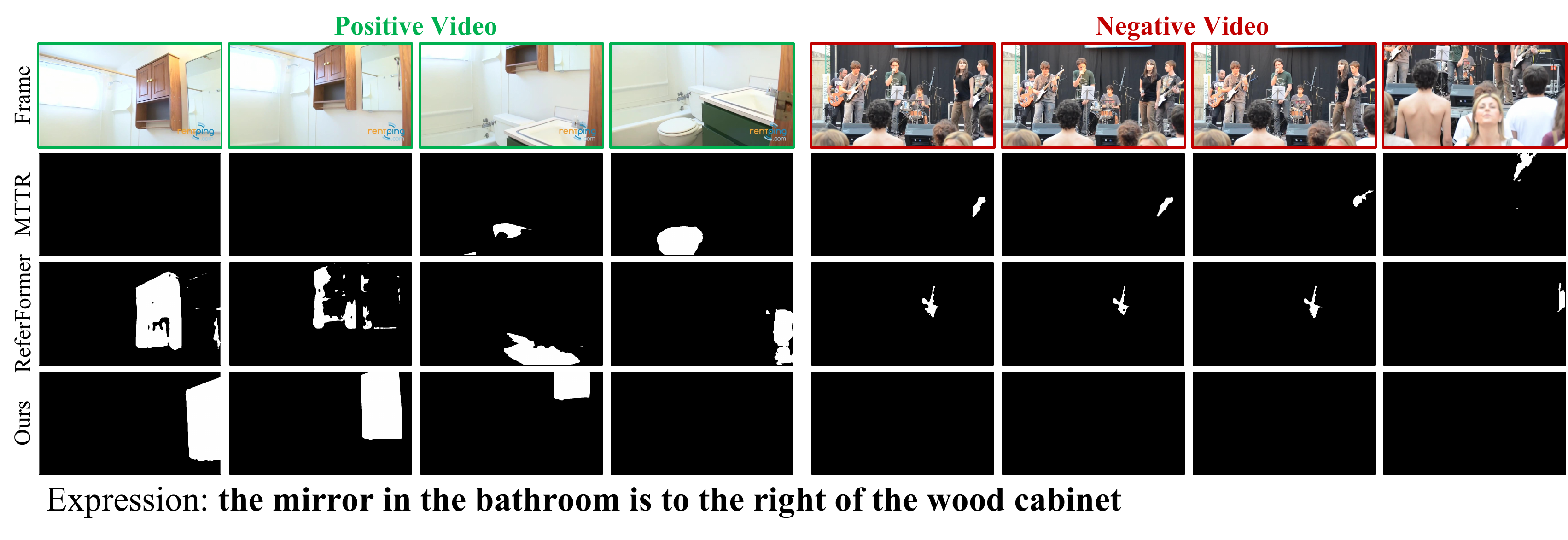}
    \vspace{-0.6cm}
    \caption{\textbf{Qualitative comparison} to the state-of-the-art R-VOS method on the $\mathrm{R}^2$-VOS task.
    }
    \vspace{-0.3cm}
    \label{fig:vis}
\end{figure*}

\subsection{Dataset and Metrics}
\vspace{0.2cm}\noindent\textbf{Dataset.}\, We conduct experiments on three datasets: Ref-Youtube-VOS \cite{seo2020urvos}, Ref-DAVIS \cite{pont2017davis} and $\mathrm{R}^2$-Youtube-VOS.
We construct a new \textbf{evaluation} dataset, $\mathrm{R}^2$-Youtube-VOS, which extends the Ref-Youtube-VOS validation set with each object containing both positive and negative linguistic queries. 
We generate diverse negative text-video pairs by either replacing object properties \{category, action, appearance, position\} in the original positive expression or randomly picking another irrelevant expression to the object. In this way, we construct the negative expressions with different granularity (each video has about 10 negative expressions). 
The positive text-video pairs are the same as in Ref-Youtube-VOS and we evaluate the segmentation quality on such positive text-video pairs.

\vspace{0.2cm}\noindent\textbf{Metrics.}\,
We inherit the commonly-used region similarity $\mathcal{J}$ and contour accuracy $\mathcal{F}$~\cite{pont2017davis} metrics for segmentation quality evaluation in R-VOS and $\mathrm{R}^2$-VOS tasks. 
Note that R-VOS evaluate $\mathcal{J}$ and $\mathcal{F}$ scores with pure positive pairs, which could be different from $\mathrm{R}^2$-VOS, because we need to additionally predict if the pairs are positive first in $\mathrm{R}^2$-VOS.
We introduce two additional metrics for robustness evaluation in $\mathrm{R}^2$-VOS tasks:
\begin{itemize}
\setlength{\itemsep}{1pt}
\setlength{\parsep}{0pt}
\setlength{\parskip}{0pt}
    \item \textbf{Semantic alignment accuracy $\mathcal{A}$.} $\mathcal{A}$ is the binary classification accuracy for semantic alignment.
    \item \textbf{Robustness score $\mathcal{R}$.} The alignment accuracy may not comprehensively represent the robustness quality, since even with false alignment, we can still get the right masks using mask confidence (For example, a negative pair is misclassified as positive, but the segmentation masks are predicted with very low confidence, and we still get the correct empty masks). 
    Therefore, we propose the new metric $\mathcal{R}=1-\frac{\sum_{\mathrm\mathbf{{M}}\in\mathcal{M}_{neg}}|\mathrm\mathbf{{M}}|}{\sum_{\mathrm\mathbf{{M}}\in\mathcal{M}_{pos}}|\mathrm\mathbf{{M}}|}$ to evaluate the degree of misclassified pixels in negative videos, where $\mathcal{M}_{neg}$ and $\mathcal{M}_{pos}$ are the sets containing segmented masks in negative and positive videos respectively. $|M|$ denotes the foreground area of mask $M$. 
    The total foreground area of positive videos $\sum_{\mathrm\mathbf{{M}}\in\mathcal{M}_{pos}}|\mathrm\mathbf{{M}}|$ serves as a normalization term.
    Ideally, a model should treat all the negative videos as backgrounds with $\mathcal{R}=1$.
\end{itemize}

\subsection{Implementation Details}
Following previous methods \cite{pminet,referformer}, our model is first pre-trained on Ref-COCO/+/g dataset \cite{yu2016refcoco,mao2016generation} and then finetuned on Ref-Youtube-VOS. The model is trained for 6 epochs with a learning rate multiplier of 0.1 at the 3rd and the 5th epoch. The initial learning rate is 1e-4 and a learning rate multiplier of 0.5 is applied to the backbone. We adopt a $\mathrm{batchsize}$ of 8 and an AdamW \cite{loshchilov2017adamw} optimizer with weight decay $1\times10^{-4}$. Following convention \cite{botach2021mttr}, the evaluation on Ref-DAVIS directly uses models trained on Ref-Youtube-VOS without re-training. All images are cropped to have the longest side of 640 pixels and the shortest side of 360 pixels during evaluation. The window size is set to 5 for all backbones.
We train model with negative samples created by randomly mismatching vision-language pairs in Ref-Youtube-VOS. We set the $\lambda_{text}=0.1$, $\lambda_{cls}=2$, $\lambda_{mask}=2$, $\lambda_{align}=1$, $\lambda_{angle}=10$, $\lambda_{L1}=5$, $\lambda_{giou}=2$, $\lambda_{dice}=2$ and $\lambda_{focal}=5$ during all training process.

\begin{table}[t]
\centering
\scalebox{0.7}{
    \begin{tabular}{l|p{0.15cm}<{\centering}|p{1.5cm}<{\centering}|p{0.7cm}<{\centering}|p{0.6cm}<{\centering}p{0.6cm}<{\centering}|p{0.7cm}<{\centering}|p{0.6cm}<{\centering}p{0.6cm}<{\centering}} 
    \hline
    \hline
    \multirow{2}*{Method} & \multirow{2}*{P} & \multirow{2}*{Backbone} & \multicolumn{3}{c|}{Ref-Youtube-VOS} &
    \multicolumn{3}{c}{Ref-DAVIS-17}\\
    \cline{4-6}\cline{7-9}
    ~ & & & $\mathcal{J} \& \mathcal{F}$ & $\mathcal{J}$ & $\mathcal{F}$ & $\mathcal{J} \& \mathcal{F}$ & $\mathcal{J}$ & $\mathcal{F}$\\
    \hline
    \multicolumn{8}{c}{Spatial Visual Backbone} \\
    \hline
    CMSA \cite{ye2019cmsa} & & ResNet50 & 34.9 & 33.3 & 36.5 & 34.7 & 32.2 & 37.2 \\
    CMSA + RNN \cite{ye2019cmsa} & & ResNet50 & 36.4 & 34.8 & 38.1 & 40.2 & 36.9 & 43.5 \\
    URVOS \cite{seo2020urvos} & & ResNet50 & 47.2 & 45.3 & 49.2 & 51.5 & 47.3 & 56.0 \\
    LBDT \cite{ding2022language} & & LBDT & 49.4 & 48.2 & 50.6 & 54.5 & - & - \\
    Wu \textit{et al.} \cite{wu2022multi} & & ResNet50 & 49.7 & 48.4 & 51.0 & 57.9 & 53.9 & 62.0 \\
    PMINet \cite{pminet} & & ResNet101 & 53.0 & 51.5 & 54.5 & - & - & - \\
    YOFO \cite{li2022you} & & ResNet50 & 54.4 & 50.1 & 58.7 & 52.2 & 47.5 & 56.8 \\
    CITD \cite{liang2021topdown} & & ResNet101 & 56.4 & 54.8 & 58.1 & - & - & - \\
    ReferFormer \cite{referformer} & \checkmark & ResNet50 & 55.6 & 54.8 & 56.5 & 58.5 & 55.8 & 61.3 \\
    \textbf{\rrvos} & \checkmark & ResNet50 & \textbf{57.3} & \textbf{56.1} & \textbf{58.4} & \textbf{59.7} & \textbf{57.2} & \textbf{62.4} \\
    \hline
    ReferFormer \cite{referformer} & \checkmark & Swin-T & 58.7 & 57.6 & 59.9 & - & - & - \\
    \textbf{\rrvos} & \checkmark & Swin-T & \textbf{60.2} & \textbf{58.9} & \textbf{61.5} & - & - & - \\
    \hline
    \multicolumn{9}{c}{Spatio-temporal Visual Backbone} \\
    \hline
    
    MTTR \cite{botach2021mttr} & & V-Swin-T & 55.3 & 54.0 & 56.6 & - & - & -\\
    Chen \textit{et al.} \cite{chen2022multi} & & V-Swin-T & 55.6 & 54.0 & 55.6 & - & - & - \\
    ReferFormer \cite{referformer} &  & V-Swin-T  & 56.0 & 54.8 & 57.3 & - & - & -\\
        \textbf{\rrvos} & & V-Swin-T & \textbf{{57.1}} & \textbf{{55.9}} & \textbf{{58.2}} &  - & - & -\\ \hline
    ReferFormer \cite{referformer} & \checkmark & V-Swin-T  & 59.4 & 58.0 & 60.9 & - & - & -\\
    \textbf{\rrvos}  & \checkmark & V-Swin-T & \textbf{61.3} & \textbf{59.6} & \textbf{63.1} & - & - & -\\
    \hline
    \hline
    \end{tabular}
}
\vspace{0.0cm}
\caption{\textbf{\textbf{Quantitative comparison} to state-of-the-art R-VOS methods.} ``P'' denotes models pretrained on ref-coco+ dataset.}
\label{tab:ytrvos}
\vspace{-0.5cm}
\end{table}

\vspace{0.2cm}
\subsection{Main Results}
\noindent\textbf{Comparison on R-VOS.} In \cref{tab:ytrvos}, we first compare our method on Ref-Youtube-VOS. For results of ResNet \cite{he2016resnet} backbone, our method achieves 57.3 $\mathcal{J}\&\mathcal{F}$ which outperforms the latest method ReferFormer \cite{referformer} by 1.7 $\mathcal{J}\&\mathcal{F}$. In addition, our method runs at 30 FPS compared to 22 FPS of state-of-the-art ReferFormer (FPS is measured using a single NVIDIA P40 with $batchsize=1$). For results of Swin-Transformer \cite{liu2021video} backbones, our method achieves 60.2 $\mathcal{J}\&\mathcal{F}$ and 61.3 $\mathcal{J}\&\mathcal{F}$ with Swin-Tiny and Video-Swin-Tiny backbones respectively, which outperforms ReferFormer \cite{referformer} and MTTR \cite{botach2021mttr} by a clear margin. Our method achieves 59.7 $\mathcal{J}\&\mathcal{F}$ on Ref-DAVIS-17 dataset, which outperforms ReferFormer by 1.2 $\mathcal{J}\&\mathcal{F}$. The performance improvement on R-VOS task indicates that the semantic consensus consolidation from cyclic training can benefit the segmentation result.

\vspace{0.2cm}\noindent\textbf{Comparison on $\mathrm{\mathbf{R}}^2$-VOS.} As shown in \cref{tab:r2vos}, we compare our method with state-of-the-art method MTTR \cite{botach2021mttr} and ReferFormer \cite{referformer} on $\mathrm{\mathbf{R}}^2$-VOS task. We extend the scoring head of MTTR and ReferFormer to discriminate semantic alignment and retrain the model with the same dataset setting as ours. We evaluate the $\mathbf{R}$-VOS task in terms of both segmentation quality and robustness quality. Generally, the segmentation quality of all methods drops compared to the result of R-VOS task. This can be accounted for two reasons. First, the positive videos can be misclassified into negative videos thus the whole video will be segmented as background. Thereby, the semantic discrimination capability will also influence the segmentation quality. Second, for MTTR and ReferFormer, the network design only focuses on positive video segmentation. For example, MTTR requires segmenting all objects in the video and then ranking them. However, by feeding unpaired inputs, the network training can be distracted. Compared to the large drop of $\mathcal{J} \& \mathcal{F}$ by MTTR and ReferFormer, the segmentation quality of our method is slightly influenced. For the robustness quality, our method achieves a 51.2 $\mathcal{R}$ and 64.5 $\mathcal{A}$ which outperforms MTTR and ReferFormer by a clear margin. 

\begin{table}[t]
    \centering
    \scalebox{0.85}{
    \begin{tabular}{l|ccc|cc}
    \hline
    \hline
    \multirow{2}*{Method} & \multicolumn{3}{c|}{Segm. Quality} & \multicolumn{2}{c}{Robust. Quality} \\
    \cline{2-4}\cline{5-6}
    ~ & $\mathcal{J}$ \& $\mathcal{F}$ & $\mathcal{J}$ & $\mathcal{F}$ & $\mathcal{R}$ & $\mathcal{A}$ \\
    \hline
    ReferFormer \cite{referformer} & 42.2 & 41.2 & 43.2 & 46.3 & 54.4 \\
    MTTR \cite{botach2021mttr} & 50.0 & 50.5 & 51.5 & 41.4 & 49.3 \\
    \textbf{\rrvos} & \bf57.2 & \bf56.0 & \bf58.4 & \bf51.2 & \bf64.5 \\
    \hline
    \hline
    \end{tabular}
    }
     \vspace{0.1cm}
    \caption{\textbf{Quantitative comparison to state-of-the-art R-VOS methods for $\mathrm{R}^2$-VOS task.}}
    \vspace{-0.5cm}
    \label{tab:r2vos}
\end{table}

\begin{figure}[t]
    \centering
    \includegraphics[width=\linewidth]{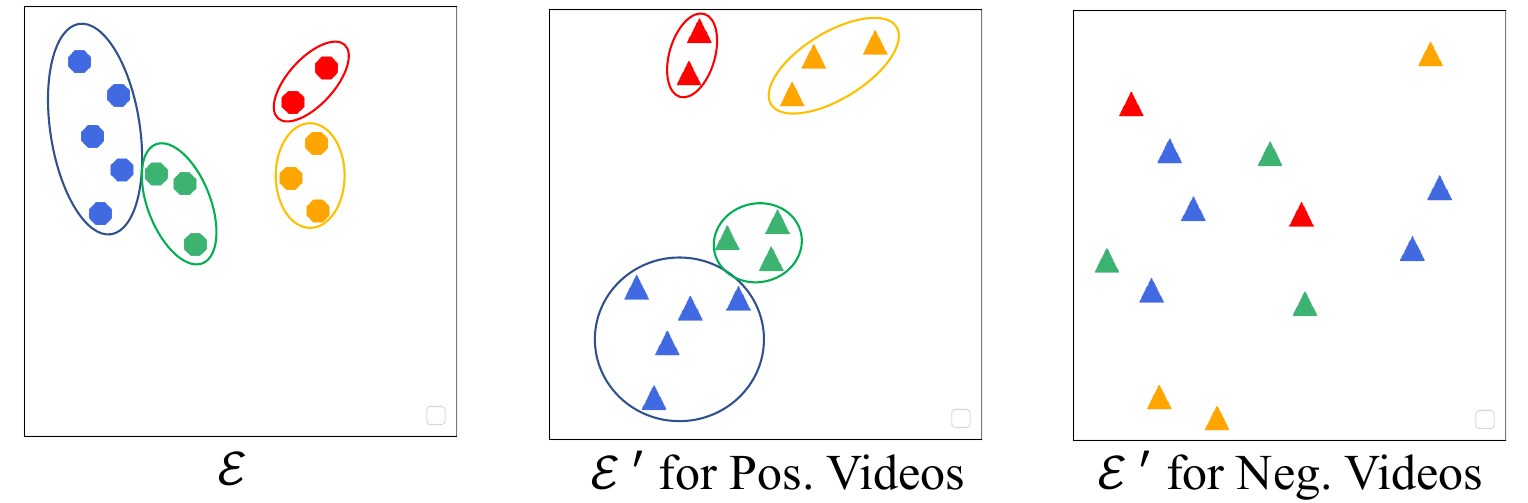}
    \vspace{-0.4cm}
    \caption{\textbf{Visualization of text embedding spaces}.
    Dots represent original text embeddings in $\mathcal{E}$, and triangles represent reconstructed ones in $\mathcal{E}'$ induced by positive (the middle sub-figure) and negative (the right sub-figure) videos.
    Elements in the same color belong to the same object. Note that an object can have multiple text descriptions.
    The structure of $\mathcal{E}'$ is well preserved from $\mathcal{E}$ for positive videos (ellipses bound embeddings of same objects), while it is not preserved for negative videos.
    }
    \vspace{-0.4cm}
    \label{fig:embed}
\end{figure}

\input{src/tables}
\vspace{0.2cm}\noindent\textbf{Qualitative results.}
We compare the qualitative results of our method against state-of-the-art methods in \cref{fig:vis} on $\mathrm{R}^2$-VOS. 
For \textbf{positive videos}: The result indicates that our method predicts accurate and temporally consistent results, while ReferFormer \cite{referformer} and MTTR \cite{botach2021mttr} fail to locate the correct object. For \textbf{negative videos}: 
Both ReferFormer and MTTR suffer from a severe false-alarm problem when the referred object does not exist in the video. 
In contrast, with multi-modal cycle constraint and consensus discrimination, our method successfully filters out negative videos and mitigates the false alarm. 
To further explore how consensus discrimination works, we visualize the text embedding and reconstructed text embedding spaces for both positive and negative videos. As shown in \cref{fig:embed}, we notice that, for embeddings of positive videos, they preserve relative relations well, while for negative videos, the reconstructed embeddings have a random pattern in the space.

\vspace{-0.1cm}
\subsection{Ablation Study}
\vspace{-0.1cm}
\label{sec:ablation}

\begin{table}[t]
    \centering
\scalebox{0.8}{
    \begin{tabular}{l|ccc|cc}
    \hline
    \hline
    Components & $\mathcal{J} \& \mathcal{F}$ & $\mathcal{J}$ & $\mathcal{F}$ & $\mathcal{R}$ & $\mathcal{A}$\\
    \hline
    \hline
    \multicolumn{6}{c}{R-VOS Task} \\
    \hline
    Baseline & 52.4 & 51.9 & 52.8 & - & - \\
    +OLM & 55.5 & 54.4 & 56.5 & - & -\\
    +OLM+CC & 56.9 & 55.7 & 58.1 & - & -\\
    +OLM+CC+FT & \bf57.3 & \textbf{56.1} & \bf58.4 & - & - \\
    \hline
    \multicolumn{6}{c}{$\mathrm{R}^2$-VOS Task} \\
    \hline
    Baseline & 41.4 & 40.8 & 42.0 & 48.5 & 53.2 \\
    +OLM+CC & 56.8 & 55.7 & 58.0 & 50.0 & 62.1\\
    +OLM+CC+FT & \bf57.2 & \bf56.0 & \textbf{58.4} & \bf51.2 & \bf64.5\\
    \hline
    \hline
    \end{tabular}}
     \vspace{0.1cm}
    \centering
    \caption{\textbf{Impact of different components in our method.} OLM: Object localizing module, CC: Consistency constraint, FT: Fusing text embeddings.}
    \label{tab:module_effectiveness}
    \vspace{-0.2cm}
\end{table}

\vspace{0.2cm}\noindent\textbf{Module effectiveness.} We conduct ablation study with the ResNet-50 backbone on Ref-Youtube-VOS and $\mathrm{R}^2$-Youtube-VOS to investigate the contribution of each component. We build a transformer-based baseline model and equip our proposed components step-by-step. As shown in \cref{tab:module_effectiveness}, we first evaluate the baseline model on R-VOS task which achieves a 52.4 $\mathcal{J}\&\mathcal{F}$.
After employing the OLM, the performance boosts to 55.5 $\mathcal{J} \& \mathcal{F}$ and the full model achieves the best performance of 57.3 $\mathcal{J} \& \mathcal{F}$. For $\mathrm{R}^2$-VOS task, since the baseline does not have a cycle consensus criterion, it shows low robustness. By equipping the consistency constraint and text embedding fusing, our method achieves the best (57.2 $\mathcal{J} \& \mathcal{F}$, 51.2 $\mathcal{R}$ and 64.5 $\mathcal{A}$).

\vspace{0.2cm}\noindent\textbf{Consistency constraint.}\,
We conduct experiments to ablate the influence of cycle consistency constraints. As shown in \cref{tab:constraint}, utilizing point-wise consistency constraint shows an inferior performance compared to the setting with structural constraint. We consider the point-wise constraint may force an injective mapping from the textual domain to the visual domain. However, the mapping can be a many-to-one function for R-VOS, \ie, each object corresponds to multiple textual descriptions. In addition, since the OLM leverages the text feature to locate the referred object, if we use the direct point-wise constraint to form reconstructed text embedding, it will encourage the network to memorize the text feature in the $\mathrm{\mathbf{f_{proxy}}}$ and result in a collapse for text reconstruction. \cref{tab:constraint} shows that the model achieves the best result of 56.8 $\mathcal{J}\&\mathcal{F}$, 50.0 $\mathcal{R}$ and 62.1$\mathcal{A}$ when with both structural angle and structural distance constraint. Note that we disable the text embedding fusion in this experiment.


\begin{table}[t]
    \centering
\scalebox{0.8}{
    \begin{tabular}{c|ccc|cc}
    \hline
    \hline
    CC & $\mathcal{J} \& \mathcal{F}$ & $\mathcal{J}$ & $\mathcal{F}$ & $\mathcal{R}$ & $\mathcal{A}$ \\
    \hline
    PW & 52.4 & 51.2 & 53.6 & 47.2 & 51.1 \\
    SA & 56.3 & 55.1 & 57.5 & 49.8 & 60.0\\
    SD & 56.0 & 55.0 & 57.0 & 47.2 & 59.3 \\
    SD+SA & \textbf{56.8} & \textbf{55.7} & \textbf{58.0} & \bf50.0 & \bf62.1\\
    \hline
    \hline
    \end{tabular}}
        \vspace{0.1cm}
    \caption{\textbf{Impact of the cycle consistency (CC) constraint.} PW: Point-wise. SA: Structural angle. SD: Structural distance.}
    \label{tab:constraint}
\end{table}

\begin{table}[t]
\vspace{-0.3cm}
\centering
    \scalebox{0.8}{
    \begin{tabular}{c|ccc|cc}
    \hline
    \hline
    Query Number & $\mathcal{J} \& \mathcal{F}$ & $\mathcal{J}$ & $\mathcal{F}$ & $\mathcal{R}$ & $\mathcal{A}$\\
    \hline
    1 & 54.9 & 54.2 & 55.6 & \bf51.9 & 63.4\\
    5 & \textbf{57.2} & 56.0 & \textbf{58.4} & 51.2 & \bf64.5 \\
    9 & 57.0 & \textbf{56.8} & 57.2 & 50.5 & 64.1 \\
    \hline
    \hline
    \end{tabular}}
        \vspace{0.1cm}
    \caption{\textbf{Impact of the query number.}}
    \vspace{-0.4cm}
    \label{tab:query_num}
\end{table}

\vspace{0.1cm}\noindent\textbf{Instance query number.}\,
Though only one object is referred to for each frame for both R-VOS and $\mathrm{R}^2$-VOS tasks, to help the network optimization, we employ more than one instance query to each video. \cref{tab:query_num} indicates that a query number of 5 brings the best overall segmentation quality.

%% file: src/6-conclusion.tex
\section{Conclusion}
In this paper, we investigate the semantic misalignment problem in R-VOS task. A pipeline jointly models the referring segmentation and text reconstruction problem, equipped with a structural cycle consistency constraint, is introduced to discriminate and enhance the semantic consensus between visual and textual modalities. For model robustness evaluation, we extend the R-VOS task to accept unpaired inputs and collect a corresponding $\mathrm{R}^2$-Youtube-VOS dataset. We observe a severe false-alarm problem suffered from previous methods on $\mathrm{R}^2$-Youtube-VOS while ours accurately discriminates unpaired inputs and segments high-quality masks for paired inputs. Our method achieves state-of-the-art performance on R-VOS benchmarks and \rrytbvos~dataset. We believe that, with unpaired inputs, \rrvos{} is a more general setting, which can shed light on a new direction to investigate the robustness of referring segmentation.


%% file: iccv_main.bbl
\begin{thebibliography}{10}\itemsep=-1pt

\bibitem{botach2021mttr}
Adam Botach, Evgenii Zheltonozhskii, and Chaim Baskin.
\newblock End-to-end referring video object segmentation with multimodal
  transformers.
\newblock In {\em Proceedings of the IEEE/CVF Conference on Computer Vision and
  Pattern Recognition}, pages 4985--4995, 2022.

\bibitem{chen2018knowledge}
Kan Chen, Jiyang Gao, and Ram Nevatia.
\newblock Knowledge aided consistency for weakly supervised phrase grounding.
\newblock In {\em Proceedings of the IEEE Conference on Computer Vision and
  Pattern Recognition}, pages 4042--4050, 2018.

\bibitem{chen2022multi}
Weidong Chen, Dexiang Hong, Yuankai Qi, Zhenjun Han, Shuhui Wang, Laiyun Qing,
  Qingming Huang, and Guorong Li.
\newblock Multi-attention network for compressed video referring object
  segmentation.
\newblock In {\em Proceedings of the 30th ACM International Conference on
  Multimedia}, pages 4416--4425, 2022.

\bibitem{chen2020dynamic}
Yinpeng Chen, Xiyang Dai, Mengchen Liu, Dongdong Chen, Lu Yuan, and Zicheng
  Liu.
\newblock Dynamic convolution: Attention over convolution kernels.
\newblock In {\em Proceedings of the IEEE/CVF Conference on Computer Vision and
  Pattern Recognition}, pages 11030--11039, 2020.

\bibitem{chen2019referring}
Yi-Wen Chen, Yi-Hsuan Tsai, Tiantian Wang, Yen-Yu Lin, and Ming-Hsuan Yang.
\newblock Referring expression object segmentation with caption-aware
  consistency.
\newblock {\em arXiv preprint arXiv:1910.04748}, 2019.

\bibitem{VLT}
Henghui Ding, Chang Liu, Suchen Wang, and Xudong Jiang.
\newblock Vision-language transformer and query generation for referring
  segmentation.
\newblock In {\em Proceedings of the IEEE/CVF International Conference on
  Computer Vision}, pages 16321--16330, 2021.

\bibitem{ding2022language}
Zihan Ding, Tianrui Hui, Junshi Huang, Xiaoming Wei, Jizhong Han, and Si Liu.
\newblock Language-bridged spatial-temporal interaction for referring video
  object segmentation.
\newblock In {\em Proceedings of the IEEE/CVF Conference on Computer Vision and
  Pattern Recognition}, pages 4964--4973, 2022.

\bibitem{pminet}
Zihan Ding, Tianrui Hui, Shaofei Huang, Si Liu, Xuan Luo, Junshi Huang, and
  Xiaoming Wei.
\newblock Progressive multimodal interaction network for referring video object
  segmentation.
\newblock {\em The 3rd Large-scale Video Object Segmentation Challenge},
  page~7, 2021.

\bibitem{he2016resnet}
Kaiming He, Xiangyu Zhang, Shaoqing Ren, and Jian Sun.
\newblock Deep residual learning for image recognition.
\newblock In {\em Proceedings of the IEEE conference on computer vision and
  pattern recognition}, pages 770--778, 2016.

\bibitem{hu2016segmentation}
Ronghang Hu, Marcus Rohrbach, and Trevor Darrell.
\newblock Segmentation from natural language expressions.
\newblock In {\em European Conference on Computer Vision}, pages 108--124.
  Springer, 2016.

\bibitem{ke2020rethinking}
Guolin Ke, Di He, and Tie-Yan Liu.
\newblock Rethinking positional encoding in language pre-training.
\newblock {\em arXiv preprint arXiv:2006.15595}, 2020.

\bibitem{kuhn1955hungarian}
Harold~W Kuhn.
\newblock The hungarian method for the assignment problem.
\newblock {\em Naval research logistics quarterly}, 2(1-2):83--97, 1955.

\bibitem{lei2018tvqa}
Jie Lei, Licheng Yu, Mohit Bansal, and Tamara~L Berg.
\newblock Tvqa: Localized, compositional video question answering.
\newblock {\em arXiv preprint arXiv:1809.01696}, 2018.

\bibitem{li2022you}
Dezhuang Li, Ruoqi Li, Lijun Wang, Yifan Wang, Jinqing Qi, Lu Zhang, Ting Liu,
  Qingquan Xu, and Huchuan Lu.
\newblock You only infer once: Cross-modal meta-transfer for referring video
  object segmentation.
\newblock In {\em AAAI Conference on Artificial Intelligence}, 2022.

\bibitem{li2021referring}
Muchen Li and Leonid Sigal.
\newblock Referring transformer: A one-step approach to multi-task visual
  grounding.
\newblock {\em Advances in Neural Information Processing Systems}, 34, 2021.

\bibitem{li2018referring}
Ruiyu Li, Kaican Li, Yi-Chun Kuo, Michelle Shu, Xiaojuan Qi, Xiaoyong Shen, and
  Jiaya Jia.
\newblock Referring image segmentation via recurrent refinement networks.
\newblock In {\em Proceedings of the IEEE Conference on Computer Vision and
  Pattern Recognition}, pages 5745--5753, 2018.

\bibitem{li2019dice}
Xiaoya Li, Xiaofei Sun, Yuxian Meng, Junjun Liang, Fei Wu, and Jiwei Li.
\newblock Dice loss for data-imbalanced nlp tasks.
\newblock {\em arXiv preprint arXiv:1911.02855}, 2019.

\bibitem{liang2021clawcranenet}
Chen Liang, Yu Wu, Yawei Luo, and Yi Yang.
\newblock Clawcranenet: Leveraging object-level relation for text-based video
  segmentation.
\newblock {\em arXiv preprint arXiv:2103.10702}, 2021.

\bibitem{liang2021topdown}
Chen Liang, Yu Wu, Tianfei Zhou, Wenguan Wang, Zongxin Yang, Yunchao Wei, and
  Yi Yang.
\newblock Rethinking cross-modal interaction from a top-down perspective for
  referring video object segmentation.
\newblock {\em arXiv preprint arXiv:2106.01061}, 2021.

\bibitem{lin2017feature}
Tsung-Yi Lin, Piotr Doll{\'a}r, Ross Girshick, Kaiming He, Bharath Hariharan,
  and Serge Belongie.
\newblock Feature pyramid networks for object detection.
\newblock In {\em Proceedings of the IEEE conference on computer vision and
  pattern recognition}, pages 2117--2125, 2017.

\bibitem{lin2017focal}
Tsung-Yi Lin, Priya Goyal, Ross Girshick, Kaiming He, and Piotr Doll{\'a}r.
\newblock Focal loss for dense object detection.
\newblock In {\em Proceedings of the IEEE international conference on computer
  vision}, pages 2980--2988, 2017.

\bibitem{lin2014microsoft}
Tsung-Yi Lin, Michael Maire, Serge Belongie, James Hays, Pietro Perona, Deva
  Ramanan, Piotr Doll{\'a}r, and C~Lawrence Zitnick.
\newblock Microsoft coco: Common objects in context.
\newblock In {\em European conference on computer vision}, pages 740--755.
  Springer, 2014.

\bibitem{CMPC}
Si Liu, Tianrui Hui, Shaofei Huang, Yunchao Wei, Bo Li, and Guanbin Li.
\newblock Cross-modal progressive comprehension for referring segmentation.
\newblock {\em IEEE Transactions on Pattern Analysis and Machine Intelligence},
  2021.

\bibitem{liu2019use}
Yang Liu, Samuel Albanie, Arsha Nagrani, and Andrew Zisserman.
\newblock Use what you have: Video retrieval using representations from
  collaborative experts.
\newblock {\em arXiv preprint arXiv:1907.13487}, 2019.

\bibitem{liu2019roberta}
Yinhan Liu, Myle Ott, Naman Goyal, Jingfei Du, Mandar Joshi, Danqi Chen, Omer
  Levy, Mike Lewis, Luke Zettlemoyer, and Veselin Stoyanov.
\newblock Roberta: A robustly optimized bert pretraining approach.
\newblock {\em arXiv preprint arXiv:1907.11692}, 2019.

\bibitem{liu2021video}
Ze Liu, Jia Ning, Yue Cao, Yixuan Wei, Zheng Zhang, Stephen Lin, and Han Hu.
\newblock Video swin transformer.
\newblock In {\em Proceedings of the IEEE/CVF Conference on Computer Vision and
  Pattern Recognition}, pages 3202--3211, 2022.

\bibitem{loshchilov2017adamw}
Ilya Loshchilov and Frank Hutter.
\newblock Decoupled weight decay regularization.
\newblock {\em arXiv preprint arXiv:1711.05101}, 2017.

\bibitem{CGAN}
Gen Luo, Yiyi Zhou, Rongrong Ji, Xiaoshuai Sun, Jinsong Su, Chia-Wen Lin, and
  Qi Tian.
\newblock Cascade grouped attention network for referring expression
  segmentation.
\newblock In {\em Proceedings of the 28th ACM International Conference on
  Multimedia}, pages 1274--1282, 2020.

\bibitem{MCN}
Gen Luo, Yiyi Zhou, Xiaoshuai Sun, Liujuan Cao, Chenglin Wu, Cheng Deng, and
  Rongrong Ji.
\newblock Multi-task collaborative network for joint referring expression
  comprehension and segmentation.
\newblock In {\em Proceedings of the IEEE/CVF Conference on computer vision and
  pattern recognition}, pages 10034--10043, 2020.

\bibitem{mao2016generation}
Junhua Mao, Jonathan Huang, Alexander Toshev, Oana Camburu, Alan~L Yuille, and
  Kevin Murphy.
\newblock Generation and comprehension of unambiguous object descriptions.
\newblock In {\em Proceedings of the IEEE conference on computer vision and
  pattern recognition}, pages 11--20, 2016.

\bibitem{miech2018learning}
Antoine Miech, Ivan Laptev, and Josef Sivic.
\newblock Learning a text-video embedding from incomplete and heterogeneous
  data.
\newblock {\em arXiv preprint arXiv:1804.02516}, 2018.

\bibitem{pont2017davis}
Jordi Pont-Tuset, Federico Perazzi, Sergi Caelles, Pablo Arbel{\'a}ez, Alex
  Sorkine-Hornung, and Luc Van~Gool.
\newblock The 2017 davis challenge on video object segmentation.
\newblock {\em arXiv preprint arXiv:1704.00675}, 2017.

\bibitem{rezatofighi2019giou}
Hamid Rezatofighi, Nathan Tsoi, JunYoung Gwak, Amir Sadeghian, Ian Reid, and
  Silvio Savarese.
\newblock Generalized intersection over union: A metric and a loss for bounding
  box regression.
\newblock In {\em Proceedings of the IEEE/CVF Conference on Computer Vision and
  Pattern Recognition}, pages 658--666, 2019.

\bibitem{seo2020urvos}
Seonguk Seo, Joon-Young Lee, and Bohyung Han.
\newblock Urvos: Unified referring video object segmentation network with a
  large-scale benchmark.
\newblock In {\em Computer Vision--ECCV 2020: 16th European Conference,
  Glasgow, UK, August 23--28, 2020, Proceedings, Part XV 16}, pages 208--223.
  Springer, 2020.

\bibitem{shi2020query}
Hengcan Shi, Hongliang Li, Qingbo Wu, and King~Ngi Ngan.
\newblock Query reconstruction network for referring expression image
  segmentation.
\newblock {\em IEEE Transactions on Multimedia}, 23:995--1007, 2020.

\bibitem{song2018explore}
Xiaomeng Song, Yucheng Shi, Xin Chen, and Yahong Han.
\newblock Explore multi-step reasoning in video question answering.
\newblock In {\em Proceedings of the 26th ACM international conference on
  Multimedia}, pages 239--247, 2018.

\bibitem{vaswani2017attention}
Ashish Vaswani, Noam Shazeer, Niki Parmar, Jakob Uszkoreit, Llion Jones,
  Aidan~N. Gomez, undefinedukasz Kaiser, and Illia Polosukhin.
\newblock Attention is all you need.
\newblock NIPS'17, page 6000–6010, Red Hook, NY, USA, 2017. Curran Associates
  Inc.

\bibitem{wang2021end}
Yuqing Wang, Zhaoliang Xu, Xinlong Wang, Chunhua Shen, Baoshan Cheng, Hao Shen,
  and Huaxia Xia.
\newblock End-to-end video instance segmentation with transformers.
\newblock In {\em Proceedings of the IEEE/CVF Conference on Computer Vision and
  Pattern Recognition}, pages 8741--8750, 2021.

\bibitem{wu2022multi}
Dongming Wu, Xingping Dong, Ling Shao, and Jianbing Shen.
\newblock Multi-level representation learning with semantic alignment for
  referring video object segmentation.
\newblock In {\em Proceedings of the IEEE/CVF Conference on Computer Vision and
  Pattern Recognition}, pages 4996--5005, 2022.

\bibitem{referformer}
Jiannan Wu, Yi Jiang, Peize Sun, Zehuan Yuan, and Ping Luo.
\newblock Language as queries for referring video object segmentation.
\newblock In {\em Proceedings of the IEEE/CVF Conference on Computer Vision and
  Pattern Recognition}, pages 4974--4984, 2022.

\bibitem{BUSNet}
Sibei Yang, Meng Xia, Guanbin Li, Hong-Yu Zhou, and Yizhou Yu.
\newblock Bottom-up shift and reasoning for referring image segmentation.
\newblock In {\em Proceedings of the IEEE/CVF Conference on Computer Vision and
  Pattern Recognition}, pages 11266--11275, 2021.

\bibitem{LAVT}
Zhao Yang, Jiaqi Wang, Yansong Tang, Kai Chen, Hengshuang Zhao, and Philip~HS
  Torr.
\newblock Lavt: Language-aware vision transformer for referring image
  segmentation.
\newblock In {\em Proceedings of the IEEE/CVF Conference on Computer Vision and
  Pattern Recognition}, pages 18155--18165, 2022.

\bibitem{ye2019cmsa}
Linwei Ye, Mrigank Rochan, Zhi Liu, and Yang Wang.
\newblock Cross-modal self-attention network for referring image segmentation.
\newblock In {\em Proceedings of the IEEE/CVF Conference on Computer Vision and
  Pattern Recognition}, pages 10502--10511, 2019.

\bibitem{yu2018mattnet}
Licheng Yu, Zhe Lin, Xiaohui Shen, Jimei Yang, Xin Lu, Mohit Bansal, and
  Tamara~L Berg.
\newblock Mattnet: Modular attention network for referring expression
  comprehension.
\newblock In {\em Proceedings of the IEEE Conference on Computer Vision and
  Pattern Recognition}, pages 1307--1315, 2018.

\bibitem{yu2016refcoco}
Licheng Yu, Patrick Poirson, Shan Yang, Alexander~C Berg, and Tamara~L Berg.
\newblock Modeling context in referring expressions.
\newblock In {\em European Conference on Computer Vision}, pages 69--85.
  Springer, 2016.

\bibitem{zhu2020deformable}
Xizhou Zhu, Weijie Su, Lewei Lu, Bin Li, Xiaogang Wang, and Jifeng Dai.
\newblock Deformable detr: Deformable transformers for end-to-end object
  detection.
\newblock {\em arXiv preprint arXiv:2010.04159}, 2020.

\end{thebibliography}
